\documentclass[final,times,authoryear,twocolumn]{elsarticle}

\usepackage{microtype}
\usepackage{graphicx}
\usepackage{booktabs}
\usepackage{hyperref}
\usepackage{xspace}
\usepackage{subcaption}

\usepackage{ycviu}
\usepackage{framed,multirow}

\usepackage{algpseudocode}
\usepackage{algorithm}
\usepackage{amsfonts}
\usepackage{dsfont}
\usepackage[tbtags]{amsmath}
\usepackage{amssymb}
\usepackage{mathtools}
\usepackage{amsthm}
\usepackage{tabularx}

\makeatletter
\DeclareRobustCommand\onedot{\futurelet\@let@token\@onedot}
\def\@onedot{\ifx\@let@token.\else.\null\fi\xspace}

\def\eg{\emph{e.g}\onedot} 
\def\ie{\emph{i.e}\onedot}

\def\etal{\emph{et al}\onedot}
\makeatother

\newcolumntype{C}{>{\centering\arraybackslash}X}

\newcommand{\indic}[1]{\mathds{1}_{#1}}

\DeclareMathOperator*{\argmax}{arg\,max}

\newcommand{\Attention}{\operatorname{Attention}}
\newcommand{\softmax}{\operatorname{softmax}}

\journal{Computer Vision and Image Understanding}

\begin{document}

\title{Patch-Based Stochastic Attention for Image Editing}

\author[1]{Nicolas \snm{Cherel}\corref{cor1}} 
\cortext[cor1]{Corresponding author: }
\ead{cherel@telecom-paris.fr}
\author[2]{Andrés \snm{Almansa}}
\author[1]{Yann \snm{Gousseau}}
\author[1]{Alasdair \snm{Newson}}

\address[1]{LTCI, Télécom Paris, Institut Polytechnique de Paris, France}
\address[2]{MAP5, CNRS \& Université Paris Cité, France}

\begin{abstract}
    Attention mechanisms have become of crucial importance in deep learning in recent years. These non-local operations, which are similar to traditional patch-based methods in image processing, complement local convolutions.
    However, computing the full attention matrix is an expensive step with heavy memory and computational loads. These limitations curb network architectures and performances, in particular for the case of high resolution images.
    We propose an efficient attention layer based on the stochastic algorithm PatchMatch, which is used for determining approximate nearest neighbors. We refer to our proposed layer as a ``Patch-based Stochastic Attention Layer'' (PSAL).
    Furthermore, we propose different approaches, based on patch aggregation, to ensure the differentiability of PSAL, thus allowing end-to-end training of any network containing our layer.
    PSAL has a small memory footprint and can therefore scale to high resolution images. It maintains this footprint without sacrificing spatial precision and globality of the nearest neighbors, which means that it can be easily inserted in any level of a deep architecture, even in shallower levels. 
    We demonstrate the usefulness of PSAL on several image editing tasks, such as image inpainting, guided image colorization, and single-image super-resolution. Our code is available at: \url{https://github.com/ncherel/psal}
\end{abstract}

 \maketitle

\section{Introduction}

Attention mechanisms~\citep{vaswani_attention_2017} have become of crucial importance in many deep learning architectures.
Originating in Natural Language Processing (NLP), where self-attention models have seen great successes~\citep{devlin_bert_2019}, these attention mechanisms have spread to other domains, and in particular, images.
The use of attention has helped deep learning introduce long range dependencies. This addresses a drawback of the commonly used convolutions, which are \emph{local} operations. Even if deeper networks and dilated convolutions can address this drawback by extending the receptive field of the network, they nevertheless fall short when non-local behaviors are important.
This happens in text processing where words referring to a same subject can be far apart, in video classification~\citep{wang_non-local_2018} when the action is changing position in time and space, or for image editing to maintain a global coherence~\citep{zhang_self-attention_2019,yu_generative_2018}.

In spite of this recent increase in popularity, the standard method to compute attention suffers from poor algorithmic complexity, scaling quadratically with the number of elements in a tensor.
This reduces the number of feasible applications to low resolution images or features.

Attention models share their underlying principles with non-local methods~\citep{buades2005non}, and are mostly used in conjunction with the highly popular transformers, as well as with image restoration and image editing networks.

Despite the success of attention models and transformers, their widespread use has been limited by their large memory footprint and computational cost. Early approaches to alleviate these shortcomings are based on either subsampling or spatial localization, which lead to undesirable artifacts in the context of image restoration and image editing.
Indeed, the latter approach (localization) is unable to model long range interactions, and the first approach (subsampling) is unable to provide pixel-precise attention maps, leading to blurred results or jagged edges where they should be smooth. In this work we aim at drastically reducing the memory and computational requirements of the attention layer, by a novel approach that avoids the artifacts commonly associated with subsampling or localization techniques.

It turns out that the attention layer is very closely linked to the problem of Nearest Neighbor (NN) search.
The softmax layer in effect biases the distribution of weights towards a handful of similar points.
In this paper, we show that when dealing with images, attention mechanisms can be efficiently estimated via an Approximate Nearest Neighbor (ANN) search. For this search, we turn to the prominent PatchMatch algorithm~\citep{barnes_patchmatch_2009}, a fast algorithm for ANN search that is especially efficient for comparing similar images. In order to overcome the computational limits of the traditional attention layer, we propose an attention layer which employs the PatchMatch method, specifically designed for the case of images, which we name Patch-Based Stochastic Attention Layer (PSAL).
PSAL has a small memory impact, scaling linearly with the input image size. As a result, it can be applied to large-size two-dimensional inputs and in particular allows us to apply the attention mechanism to high resolution images or to both shallow and deep 2D feature maps. Such situations are out of reach for classical attention mechanisms, because of memory limitations, and require the use of a sub-sampling strategy or a restriction to very deep features. These approaches are problematic for image editing in several situations: handling high resolution images, using low level features closer to the original image or applying attention mechanisms at the pixel level.

We illustrate the usefulness of PSAL on several image editing tasks: image colorization, image inpainting, and single-image super-resolution.
In the inpainting case in particular, we show that our approach can handle high resolution images without impairing the quality of results.

The paper is organised as follows. In Section~\ref{sec:relatedWork}, we detail previous work related to attention layers and patch-based image editing. In Section~\ref{sec:attentionModels}, we describe the classical attention layer and then the approach proposed in this paper.
In Section~\ref{sec:results}, we evaluate our approach on five tasks. Firstly, we compare theoretical and true memory usage, then we validate the approximation performance of our approach on an image reconstruction task. Thirdly, we compare PSAL with other attention layers on a task of image colorization. Fourthly, we show how PSAL can be used for image inpainting, allowing us to process high-resolution images. Finally, we show PSAL can also work in the context of single-image super-resolution.
To summarize, we propose a new attention layer for images that has a greatly reduced memory complexity compared to traditional attention layers, while maintaining a similar functionality. In particular, this means that attention-based architectures can be easily modified to process images with a much higher resolution than is currently attainable. 
As a concrete example, Full Attention memory requirements \emph{scale quadratically} with the number of pixels (requiring 16 GB of memory for a $256 \times 256$ image and an infeasible 256 GB for $512 \times 512$ images). PSAL, on the other hand, \emph{scales linearly}, and needs only 786 kB and 3.15 MB, respectively, demonstrating the great reduction of memory requirements which the proposed method entails.

\section{Related work}
\label{sec:relatedWork}

\subsection{Attention models}

The origins of attention models can be found in the work of \citet{vaswani_attention_2017} on NLP, where they became a crucial element of subsequent networks~\citep{devlin_bert_2019}. In computer vision, self-attention has been applied successfully to generative adversarial networks ~\citep{zhang_self-attention_2019}, object detection~\citep{carion_end--end_2020} and video classification~\citep{wang_non-local_2018}. Attention models are also closely related to the Non-Local operations presented by \citep{wang_non-local_2018}. In their framework, self-attention with a softmax activation function becomes a specific non-local layer.

Self-Attention has a quadratic complexity in memory $\mathcal{O}(n^2)$, which has inspired works on more efficient alternatives.
Some works have focused on reducing the amount of distances to compute.
\citet{child_generating_2019} propose the Sparse Transformer which factorizes the attention matrix with strided and fixed patterns. \citet{zaheer2020bigbird} combine random attention, local attention, and global attention.
\citet{kitaev_reformer_2020} introduce a layer based on Locality Sensitive Hashing (LSH), an ANN method.
Other methods have looked at approximations of the softmax using linear operations.
\citet{katharopoulos_et_al_2020} approximate the softmax with a linear operation through a fixed projection, making the attention much faster and more efficient.
Performers~\citep{choromanski_rethinking_2020} estimate the attention matrix with orthogonal random features.
\citet{wang2020linformer} proposed the Linformer, which factorizes the self-attention into low-rank matrices. 
\citet{tay_efficient_2020} have written a comprehensive survey of efficient transformers.

Efficient attention models for vision-related tasks are needed given the high dimensionality of images.
Local Attention~\citep{parmar_image_2018,Parmar2019} reduces the attention to a smaller neighborhood.
To further extend the attention range, Swin~\citep{liuSwinTransformerHierarchical} proposed local windows with cycling shifts.
\citet{calian_scram_2019}, close in spirit to the proposal of this paper, propose an attention layer derived from PatchMatch to compute the attention, however in their preprint the authors do not verify the validity of their layer in a practical deep learning setting, meaning that there is no way of knowing if the layer functions correctly.

Since we will be evaluating PSAL on image editing, in particular on the inpainting problem, and given the close link of this domain with patch-based methods, we now briefly introduce the context of image editing.

\subsection{Image editing}

Image editing has long used patch-based approaches for inpainting~\citep{criminisi_region_2004,wexler_space-time_2007}, retargeting~\citep{barnes_patchmatch_2009}, style transfer~\citep{frigo2016cvpr}, and other image editing tasks~\citep{darabi_image_2012}, making heavy use of NN patches. Self-similarity has also been identified as a key component for single-image super-resolution~\citep{glasnerSuperresolutionSingleImage2009,michaeliBlindDeblurringUsing2014}.
The search for NN patches, the bottleneck of such approaches, was greatly accelerated by PatchMatch~\citep{barnes_patchmatch_2009}, an ANN algorithm that works by propagating good matches to neighbors.

Recently, deep learning alternatives have used encoder-decoder architectures and the GAN framework for image inpainting~\citep{pathak_context_2016,iizuka_globally_2017}.
\citet{yu_generative_2018} have observed that the textures often lack details, and propose to use an attention layer to reuse existing patches, bridging the gap between deep learning and patch-based methods.
ShiftNet~\citep{ferrari_shift-net_2018} shares a similar idea but replaces the softmax stage by an argmax, effectively reducing the attention layer to an NN layer.

For single-image super-resolution, deep networks have progressively replaced these methods with the introduction of large datasets~\citep{zhangImageSuperResolutionUsing2018}.
Attention layers have been integrated within recent image restoration networks, increasing their modelling power with non-local operations~\citep{liu_non-local_2018,daiSecondOrderAttentionNetwork2019}. The work of \citet{meiImageSuperResolutionCrossScale2020} highlights the importance of attention across multiple scales.

Finally, diffusion models~\citep{hoDenoisingDiffusionProbabilistic2020b} have been shown to be very powerful models using attention layers for image generation or image inpainting~\citep{rombachHighResolutionImageSynthesis2022}.

\section{Patch-Based Stochastic Attention}
\label{sec:attentionModels}
\subsection{Full Attention}
\label{subsec:fullAttention}

We first give the mathematical definition of what we refer to as the \emph{Full Attention layer} (FA layer), the classical dot-product attention mechanism introduced in~\citep{vaswani_attention_2017}. Let $Q\in\mathbb{R}^{m\times d}$ denote a set of $m$ \emph{queries} packed into a matrix, each query being a vector of $\mathbb{R}^d$. Intuitively, these queries correspond to the different elements for which we want an attention vector. In the context of images, this may be a set of patches. The queries are compared to a set of $n$ keys, packed into the matrix $K \in \mathbb{R}^{n \times d}$. These keys correspond to elements that we want to use as a reference to give more or less importance to the queries. In the image case, the keys may be a set of patches (not necessarily the same as the queries). Given a vector $V \in \mathbb{R}^{n \times d'}$, the attention output is the following weighted sum:
\begin{equation}
\Attention(Q,K,V) = \softmax(QK^T)V
\label{eq:fullAttention}
\end{equation}
where the softmax function is applied to each row of the matrix $QK^T$. Recall that the softmax of a vector $x \in \mathbb{R}^n$ is defined by $\softmax(x)_i = e^{x_i} / \sum_{j=1}^n e^{x_j}$.
Thus, the final result of the attention is a vector for each query, containing a weighted average of values, weighted by the dot product of the query with the elements in $K$ (the keys).
In this paper, we concentrate on the case where these contain image patches. For simplicity, in all that follows, we will consider that $m=n$. However, we note that our approach is equally applicable in the general case where $m\neq n$.

The dot-product attention in Equation~\eqref{eq:fullAttention} requires the computation of the full matrix $QK^T$ with $n^2$ entries. This results in a computational complexity of $\mathcal{O}(n^2d)$, and a memory complexity of $\mathcal{O}(n^2)$, $n$ being the input size \eg sequence length or number of pixels. For 1-dimensional vectors, this can be implemented with simple matrix multiplications. For 2-dimensional vectors \ie~patches, dot products can be computed as 2D convolutions as remarked by \citet{li_combining_2016}.

This memory requirement is the most problematic limitation of the FA layers.
To address this problem, 
\citet{yu_generative_2018} subsample the set of keys.
Another approach, proposed by 
\citet{liu_non-local_2018} for image restoration, is to compute the attention restricted to a local neighborhood.
While such approximations are useful, they nevertheless rely on many pairwise distances to be computed.
In practice, we remark that after the softmax operation in Equation~\eqref{eq:fullAttention}, \emph{only a few elements actually matter}.
One way of viewing the attention layer is as a ``soft'' NN search layer.
Consequently, in order to limit algorithmic complexity, we propose to switch to a \emph{sparse} layer, keeping only a single non-zero value in each row in the matrix $QK^T$, corresponding to the NN. The crux of the problem is now to solve the NN search quickly and with little memory overhead.
For this purpose, we propose to employ an efficient ANN algorithm, designed specifically for images: the PatchMatch algorithm~\citep{barnes_patchmatch_2009}.

\subsection{Patch-based Stochastic Attention Layer (PSAL)}

\begin{figure*}\centering
    \includegraphics[width=0.8\textwidth]{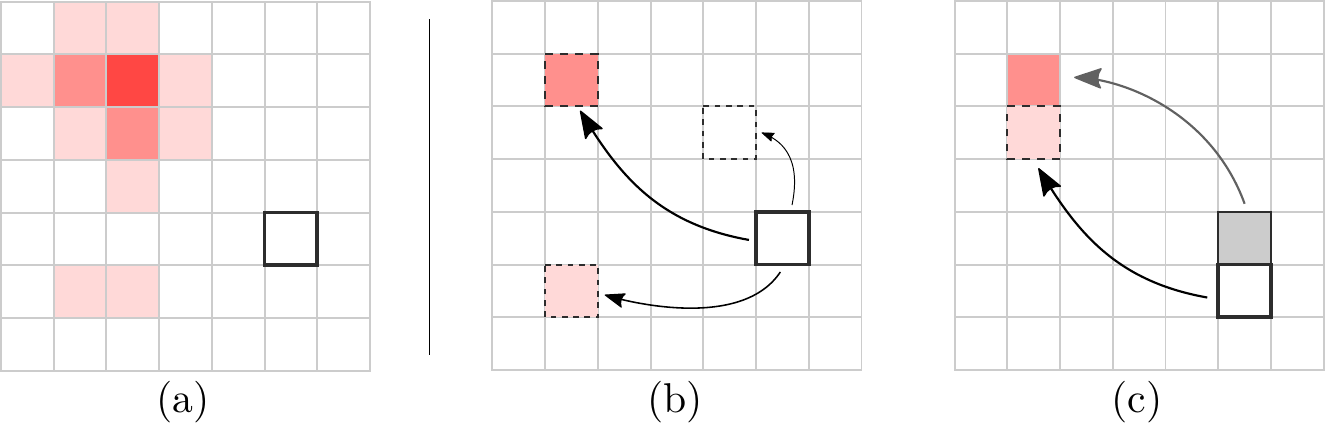}
    \caption{Illustration of patch NN search. (a) Full Attention computes a complete attention matrix but many elements have negligible weight. (b) Patch-Based Stochastic Attention only probes randomly a few elements. (c) Good matches are propagated to neighbors }
    \label{fig:patchmatch}
\end{figure*}

Recall that our goal is to replace the traditional attention layer, which is cumbersome in terms of memory, with a more efficient approach, designed for images. As we have noted above, attention layers are closely related to the search of NNs. We start by defining the NN mapping $\psi$, between the query vectors and the key vectors:
\begin{equation}
\psi(i) = \argmax_{j \in \{1,\dots, n\}} \left< Q_i ,  K_j \right>,
\label{eq:psi}
\end{equation}
where $Q_i$ is the $i$th line of the matrix $Q$, corresponding to the vector $i$, and likewise for $K$. In the attention literature, the dot product is commonly used to compare patches, but for more generality, we introduce a patch similarity function $s(Q_i,K_j)$, which is high when patches are similar.

Finally, we introduce the associated sparse matrix $A \in \mathbb{R}^{n\times n}$ defined as:
\begin{equation}
A_{i,j} = \begin{cases}
            1 \quad \mathrm{if} \quad \psi(i) = j
            \\
            0 \quad \mathrm{otherwise}
            \end{cases}.
\label{eq:matA}
\end{equation}
Our definition of attention, which can be seen as a rewriting of Equation~\eqref{eq:fullAttention}, is simply
\begin{equation}
\Attention(Q,K,V) = AV.
\label{eq:nnAttention}
\end{equation}
The next step is to provide a fast and light way to approximate $\psi$. In the general case, this can be implemented using ANN algorithms, like kd-trees or Locality Sensitive Hashing as done by \citet{kitaev_reformer_2020}. For images and image-like tensors such as feature maps, where vectors are patches, PatchMatch is an efficient alternative. It accelerates the search for NNs by drawing on a specific regularity property of images: the \emph{shift map} (ie. the values $\psi(i)-i$) between NNs in different images is approximately piece-wise constant. Note that this implicitly requires a spatial organisation (1D, 2D, etc) of the data, which is the present case of images. 

PatchMatch is an efficient, stochastic, algorithm for searching for ANNs of patches in images and videos, between a query image and a key image.
PatchMatch starts out by randomly associating ANNs to the query patches. In general, these ANNs will be of poor quality, however from time to time a good association $\psi(i)$ will be found. The algorithm then attempts to propagate the shift $\psi(i)-i$ given by this ANN to other query patches in the spatial patch neighborhood of $i$, with the hypothesis that these shifts are piece-wise constant. This happens, for example, when a coherent object is found in both the query image and the key image. 

After the random initialization, the PatchMatch algorithm relies on two alternating steps:
\begin{enumerate}
    \item Propagation, in which good shifts are propagated to spatial neighbors
    \item Random search for better ANNs for each query patch
\end{enumerate}
The random search is carried out by randomly looking in a window of decreasing size, around the current ANN. An illustration of the idea of PatchMatch can be seen in Figure~\ref{fig:patchmatch}.

A drawback of PatchMatch is that it is an iterative algorithm that is not naturally parallelizable, with the propagation step  being inherently sequential, thus making it problematic for use in deep learning.
However, we employ a semi-parallel approximation to this propagation step known as jump-flooding~\citep{barnes_patchmatch_2009,rong_jump_2006} described in Algorithm~\ref{alg:jumpflood}. A significant advantage of PatchMatch is that it keeps only the current ANN, which vastly reduces the memory requirements.

\algblock{ParFor}{EndParFor}
\algnewcommand\algorithmicparfor{\textbf{parfor}}
\algnewcommand\algorithmicpardo{\textbf{do}}
\algnewcommand\algorithmicendparfor{\textbf{end\ parfor}}
\algrenewtext{ParFor}[1]{\algorithmicparfor\ #1\ \algorithmicpardo}
\algrenewtext{EndParFor}{\algorithmicendparfor}

\begin{algorithm}[t]
    \caption{Propagation step
    using Jump-Flooding}
    \begin{algorithmic}[1]
    \Require{queries $Q$, keys $K$, ANN field $\psi$
    }
    \Ensure{Updated ANN field $\psi$}
    \ParFor{$p = 1 \dots m$}
    \For{$l = 1,2,4,8$}
        \For{$\delta$ in $[(-l,0), (l,0), (0,-l), (0,l)]$} \Comment{Look up, down, left, right}
            \State $q \gets \psi(p + \delta) - \delta$ \Comment{Candidate position $q$}
            \If{$s(Q_p, K_q) > s(Q_p, K_{\psi(p)})$}
                \State $\psi(p) \gets q$
            \EndIf
        \EndFor
    \EndFor
    \EndParFor
    \end{algorithmic}
    \label{alg:jumpflood}
\end{algorithm}

\subsection{Complexity}

The computational complexity of our proposed PSAL is $\mathcal{O}(ndN\log(n))$ with $N$ the number of iterations of propagation / random search.
The derivations are the following: the PatchMatch algorithm consists of $N$ iterations of propagation and random search for each of the $n$ patches.
Following the usual practice for random search, we sample 1 candidate in windows of geometrically decreasing size of ratio 0.5, leading to $\mathcal{O}(\log(n))$ candidates.
For the propagation step, the parallel approach (Algorithm~\ref{alg:jumpflood}) uses $4 \times 4 = 16$ candidates.
Finally, the cost of a similarity computation (dot product, cosine similarity, or L2 distance) between two patches has a complexity $\mathcal{O}(d)$ for $d$ the dimension of the patch.
Putting all together, we have a computational complexity of $\mathcal{O}(ndN \max (\log(n), 16))$.
The memory complexity is simply $\mathcal{O}(n)$, as we only have to store the shift map $\psi$, our mapping for each patch. This is to be compared with the Full Attention layer, whose complexities are  $\mathcal{O}(n^2d)$ and $\mathcal{O}(n^2)$, respectively.

In particular, the memory complexity is linear with respect to the number of queries, while that of Full Attention is quadratic. 
This has important consequences, in particular on the maximum resolution of images that can be processed by deep learning architectures which employ attention layers.

\begin{figure*}\centering
\begin{subfigure}[t]{0.45\textwidth}
    \includegraphics[width=\textwidth]{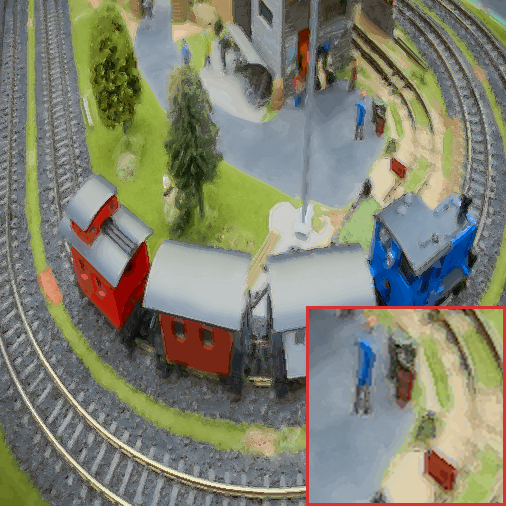}
    \caption{Full Attention}
\end{subfigure}
\hfill
\begin{subfigure}[t]{0.45\textwidth}
    \includegraphics[width=\textwidth]{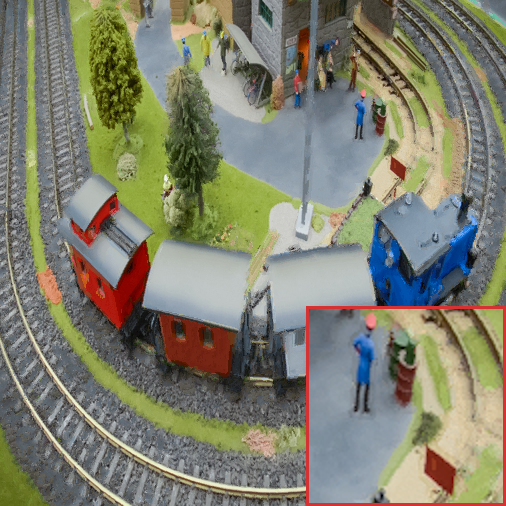}
    \caption{PSAL}
\end{subfigure}
\caption{Full Attention (left) and PSAL (right) reconstruction using another frame of the same video. The memory constraints and the subsequent subsampling step make it impossible for the Full Attention to capture all details}
\label{fig:reconstruction}
\end{figure*}

\subsection{Differentiability}
\label{subsec:differentiability}
Unfortunately PatchMatch, using a single NN, is not differentiable with respect to $Q$ and $K$ because of the argmax operator in equation~\eqref{eq:psi}.

We propose to overcome this limitation by the use of multiple neighbors, and the adaptation of the softmax operator. Note that another work, by~\citep{Plotz2018}, also proposed the use of nearest neighbours. However, it is based on a continuous relaxation of the k-Nearest Neighbor (KNN)  attention matrix, and thus has the same computational and memory complexity as Full Attention, which is precisely what we wish to avoid.

We now assume that we have access to multiple NNs instead of only one so that the support of our rows, or the shift-map, is now a set $\psi(i)$ with $k$ elements. The matrix $A$ writes:

\begin{equation}
A_{i,j} = \begin{cases}
            \frac{\exp(S_{ij})}{\sum_{j' \in \psi(i)} \exp(S_{ij'})} \quad \mathrm{if} \quad j \in \psi(i)
            \\
            0 \quad \mathrm{otherwise}
            \end{cases}.
\label{eq:matA}
\end{equation}

The matrix $S \in \mathbb{R}^{n \times n}$ is an intermediate matrix storing the similarities between elements of $Q$ and $K$. We will give further details on how $S$ is established later in the paper.

We propose two ways to construct $A$ in a computationally- and memory-efficient way: one based on using several NNs, and one based on patch aggregation.
Our theoretical findings are confirmed by the experiments (Section~\ref{sec:colorizationDifferentiability}), in which the importance of the differentiability property will be shown.

We show that this restores differentiability with respect to $Q$ and $K$ as long as the number $k = \lvert \psi(i) \rvert$ of NNs is larger than 1.

\subsubsection{Derivatives}
\label{sec:gradients}

We give the derivatives with respect to all input variables in the case of scalar values \ie $Q \in \mathbb{R}^{n \times 1}, K \in \mathbb{R}^{n \times 1}, V \in \mathbb{R}^{n \times 1}$. We have : $\forall i, j$:

\begin{equation}
    \frac{\partial (AV)_i}{\partial V_j} = A_{ij}
\end{equation}

\begin{equation}
    \frac{\partial (AV)_i}{\partial Q_j} = \indic{i=j} \sum_{k \in \psi(i)} A_{ik}  V_k \left[ \frac{\partial S_{ik}}{\partial Q_j} - \sum_{j' \in \psi(i)} A_{ij'} \frac{ \partial S_{ij'}}{\partial Q_j} \right]
\end{equation}

\begin{equation}
    \frac{\partial (AV)_i}{\partial K_j} = \indic{j \in \psi(i)} \frac{\partial S_{ij}}{\partial K_j} \sum_{k \in \psi(i)} A_{ik} \left[ \indic{j=k}  - A_{ik} \right] V_k 
\end{equation}

The derivatives with respect to $Q, K$ involve the derivatives of the similarity function used $S_{ij} = s(Q_i, K_j)$ with respect to its inputs but are of no difficulty when $s$ is the Euclidean distance or the dot product.
In the case where $\lvert \psi(i) \rvert = 1$, the derivatives with respect to $Q_j$ and $K_j$ are equal to $0$ for all $j$, confirming the need for multiple neighbors in our method.

\subsubsection{$k$-Nearest Neighbors differentiability}
In order to overcome the aforementioned differentiability problem which arises when we have only a single NN, we propose to enrich the matrix $S$ using several NNs for each patch, which can be done with a modified version of PatchMatch \citep{Barnes2010}. In this case, each $\psi(i)$ is now a \emph{set} of $k$ NN correspondences. We start by redefining the matrix $S_{i,j}$ as: 

\begin{equation}
 S_{i,j} = \begin{cases}
                s(Q_i, K_{j}) & \mathrm{if} \:  j \in \psi(i)\\
                -\infty & \text{otherwise}.
            \end{cases}
\end{equation}
Then by applying a softmax operation along the rows, we obtain the differentiable $\tilde{A}_t = \softmax_t(S)$ with the same behavior than the original implementation. It can be seen that the Full Attention implementation corresponds to $\tilde{A}_t$ in the case where $k=N$.
There is only a light computational overhead to manage the $k$ nearest neighbors using a max heap and the associated memory extension. Finally PSAL-$k$ has $\mathcal{O}(ndN\log(n)\log(k))$ computational complexity and $\mathcal{O}(nk)$ memory complexity.

\begin{table*}
    \centering
    \caption{Memory (mem.) (GB) required by the attention layer when the input size is increasing. $n$ is the number of pixels. For Local Attention, the window size $w$ is set to $50$. For Performer, we use the recommended parameter $M=d\log d$. Tested in conditions with patch size $p=7$ and 16 channels. PSAL and PSAL Aggreg. have a low enough memory footprint to make batches larger than 1 possible. $^*$ Linear Attention, Performer, and Reformer use 1D vectors, gathering patches and unrolling them consumes a significant amount of memory.}
    \begin{small}
    \begin{tabularx}{\linewidth}{|X|C|C|C|C|C|}
        \hline
        Attention Method & Mem. complexity & Mem. (GB) for 64x64 & Mem. (GB) for 128x128 & Mem. (GB) for 256x256 & Mem. (GB) for 512x512 \\
        \hline
         Full Attention & $\mathcal{O}(n^2)$ & 0.30 & 4.98 & 15.26 & 250.04 \\
         Local Attention & $\mathcal{O}(w^2n)$ & 0.19 & 0.64 & 3.23 & 13.12 \\
         Linear Attention & $\mathcal{O}(n)^*$ & 0.11 & 0.46 & 1.85 & 7.47 \\
         Performer & $\mathcal{O}(nd\log d)^*$ & 0.71 & 2.84 & 11.55 & 17.68 \\
         Reformer & $\mathcal{O}(128n)^*$ & 0.47 & 1.86 & 7.42 & 21.47 \\
         PSAL 3 & $\mathcal{O}(3n)$ & 0.01 & 0.01 & 0.04 & 0.18 \\
         PSAL Aggreg. & $\mathcal{O}(p^2n)$ & 0.05 & 0.19 & 0.74 & 2.95 \\
        \hline
    \end{tabularx}
    \end{small}
    \label{tab:benchmark}
\end{table*}

\subsubsection{Patch aggregation differentiability}

The second approach we propose is to perform spatial aggregation. Intuitively, we enrich the list of NNs for a given patch by using the NNs of the \emph{spatial} neighbors of this patch. To put it more colloquially, the neighbor of my spatial neighbor is my neighbor. In this case, we redefine $S$ in terms of a spatial neighbor $i'$ and its patch-space neighbor $j'$ as\footnote{Note that this also impacts the definition of $A$ in Equation~\eqref{eq:matA}}:

\begin{equation}
 S_{i,j} = \begin{cases}
        s(Q_{i'}, K_{j'}) &
         \text{if} \left\lbrace\begin{aligned} & \text{$i'\in\mathcal{N}_i$ and $j'\in\psi(i')$} \\ &
        \text{and $i'-i=j'-j$}
        \end{aligned}\right.\\
        -\infty & \text{otherwise}.
    \end{cases}
\label{eq:aggDiff}
\end{equation}

where $\mathcal{N}_i$ is the spatial patch neighborhood of $i$. The condition in Equation~\eqref{eq:aggDiff} basically says that, for a patch $i$, we are analyzing its \emph{spatial} neighbor $i'$ and the NN of $i'$, $\psi(i')$. We then check that the spatial shift between the patch $i$ and $i'$ is the same as the NN patches $j$ and $j'$. The last condition is necessary to link $j$ to $j'$.
In practice, we choose the spatial neighborhood to be 
 the patch neighborhood. This aggregation can be useful to other sparse attention layers. PSAL Agg. has $\mathcal{O}(p^2n)$ memory complexity and $\mathcal{O}(p^2n + ndN\log(n))$ computational complexity for a patch size of $p$.

These previous two differentiability methods can be combined as desired. We present experiments in the next Section that show that without these approaches, networks have great difficulty learning, and produce poor results.

\subsection{Similarity function}

One of the key components of the attention mechanism is the similarity function used to build the similarity matrix $S$ .
The original work of \citet{vaswani_attention_2017} used the dot product with a scaling parameter, however other options such as the cosine similarity \citep{yu_generative_2018} are better suited in some situations.
PSAL can be used with any similarity metric.
This is important for providing a flexible attention layer, especially given than the similarity function can significatively impact the performance of a whole network. For instance, we found out that the dot product similarity performs poorly compared to the $\ell_2$ distance in the reconstruction and colorization tasks (Sections~\ref{subsec:image_reconstruction} and~\ref{sec:colorization}).
Linear Attention~\citep{katharopoulos_et_al_2020} and Reformer~\citep{choromanski_rethinking_2020} are linear approximations and are thus limited in some cases.

\section{Results}
\label{sec:results}

We now present quantitative and qualitative results showing the advantages of our proposed attention layer.
We compare our layer in 5 different situations:

\begin{enumerate}
    \item We analyze the memory consumption of different attention layers. We observe in particular that PSAL requires orders of magnitude less memory than alternatives
    \item Image reconstruction. We show that by replacing the FA layer with the proposed PSAL, the NN patches reconstruct an image well. This evaluation is motivated by the fact that the examples in the matrix $V$ should reconstruct or approximate the initial queries
    \item Image colorization. PSAL performs better than other attention layers in the context of guided image colorization, an image editing task for which attention is crucial
    \item Image inpainting. We show that it is possible to replace a classical FA layer directly with a PSAL, without affecting the inpainting quality, allowing for inpainting of high-resolution images
    \item Single-image super-resolution. Similarly to inpainting, classical FA layer can be replaced with PSAL without affecting the performance but allowing high resolution processing which is beneficial
\end{enumerate}

We compare our work with Full Attention and other state-of-the-art attention layers: Local Attention~\citep{parmar_image_2018}, Performer~\citep{choromanski_rethinking_2020}, Reformer~\citep{kitaev_reformer_2020}, and Linear Attention~\citep{katharopoulos_et_al_2020}. We compare these with two differentiable PSAL approaches: PSAL 3 which uses $k$-NNs with $k=3$ and PSAL with aggregation, (PSAL Aggreg.).

\subsection{Memory benchmark}
\label{subsec:benchmark}

We recall that one of our initial motivations is to develop an efficient attention layer that can be used in any situation.
In particular, we designed our layer to not be memory-bounded when applying it to mid to large images.
In this section, we measure the memory footprint of various attention layers for different feature maps size. We also compare the theoretical complexity with the true memory occupation. These results can be seen in Table~\ref{tab:benchmark}. We see that our PSAL requires vastly less memory than FA and competing methods. For example, in the case of $512\times 512$ images, FA requires 250GB, whereas PSAL 0.18GB or 2.95GB (PSAL 3 or PSAL Aggreg.).

\begin{figure}\centering
\includegraphics[width=0.48\textwidth]{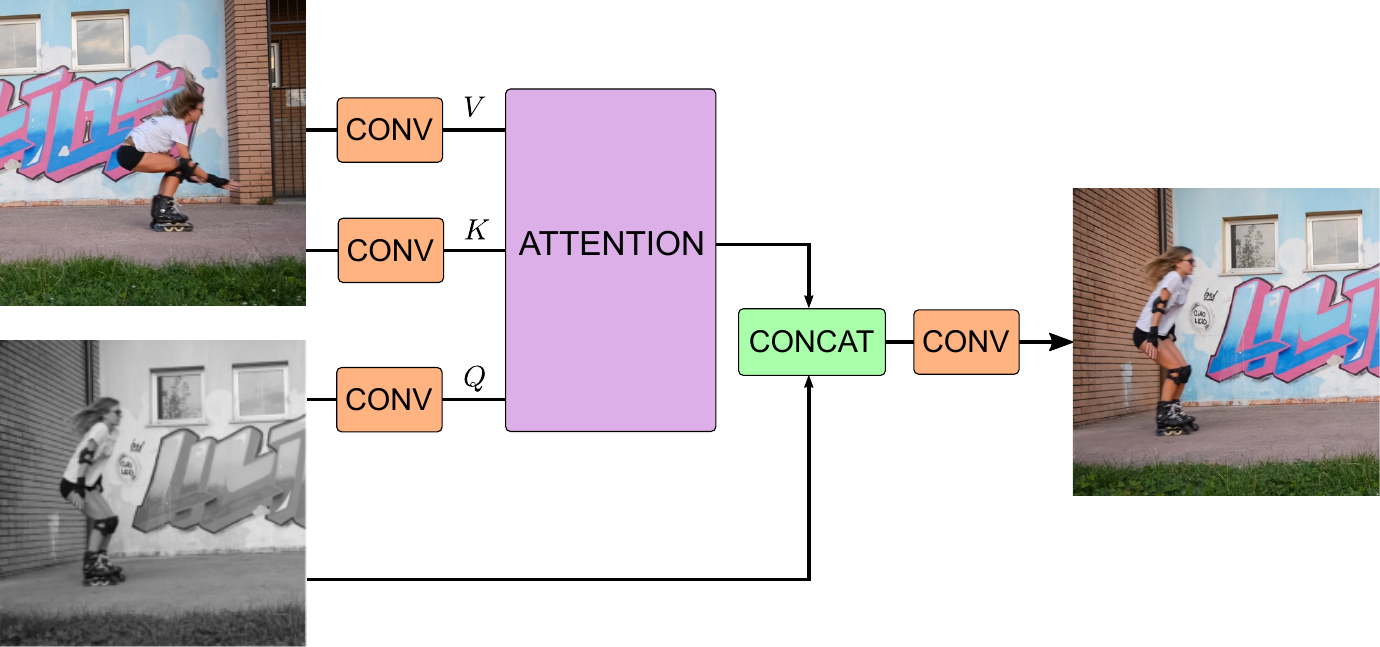}
\caption{Architecture for our colorization network. We have used as simple an architecture as possible to isolate the contribution of the attention layer.}
\label{fig:colorization}
\end{figure}

\subsection{Image reconstruction task}
\label{subsec:image_reconstruction}

We now compare the performances of PSAL and that of a FA layer on the task of image reconstruction using patches. The goal of this experiment is to check that using ANNs does not induce any loss of quality 
with respect to the ``standard'' attention. 
In the case of images, we can directly view this quality, contrary to the case of deep features.

We reconstruct an image by applying the attention layer using the $\ell_2$ distance, $Q$ the set of patches from image A, $K$ the set of patches from image B and $V$ the pixels of image B. Image A and B are taken from the same video sequence.

\begin{table}
    \centering
    \caption{Reconstruction error of PSAL layers vs other attention layers.}
    \begin{tabularx}{\linewidth}{|C|C|}
        \hline
        Attention Method & $\ell_2$ loss \\
        \hline
         Full Attention & 0.0022 \\
         Local Attention & 0.0119 \\
         Linear Attention & 0.0576 \\
         PSAL 3 & 0.0011 \\
         PSAL Aggreg. & 0.0008 \\
        \hline
    \end{tabularx}
    \label{tab:reconstruction}
\end{table}

We first show that PSAL can efficiently appproximate Full Attention on this reconstruction task, using patches of size $7 \times 7$. Table~\ref{tab:reconstruction_stride} indicates that the performances of PSAL and Full Attention (with stride 1) are equivalent but limits the size of the input to $64 \times 64$.
As the resolution is increased, an approximation has to be used for Full Attention. Among the different options, we decide to keep all the queries $Q$ but the keys $K$ are obtained using a stride to decrease the memory pressure.
We observe better results for PSAL in these cases.

We then consider a realistic case where images are of size $512\times512$.
Quantitative results for 30 pairs of images 10 frames apart are shown in Table~\ref{tab:reconstruction}. The metrics confirm that the naive approximation of Full Attention via subsampling is more damageable to the performance than the approximation of PSAL.
For Linear Attention, the approximation proposed by \citet{katharopoulos_et_al_2020}
is an approximation of the dot product which is not well-suited in this case.
Similarly Local Attention performs poorly as the displacement between the reference and the image to reconstruct is often larger than the local window.

\begin{table}
    \centering
    \caption{Reconstruction error of PSAL layers vs Full Attention.}
    \begin{tabularx}{\linewidth}{|C|C|C|C|}
        \hline
        Image size & Stride & Full Attention & PSAL 3 \\
        \hline
        512 & 10 & 0.0022 & 0.0011 \\
        256 & 5 & 0.0031 & 0.0017 \\
        128 & 2 & 0.0038 & 0.0029 \\
        64 & 1 & 0.0048 & 0.0048 \\
        \hline
    \end{tabularx}
    \label{tab:reconstruction_stride}
\end{table}

\begin{table}
    \centering
    \caption{Colorization error of PSAL layers with different parameters. PSAL-1 is not fully differentiable which limits learning and performance. These experiments confirm that the solutions we proposed in Section~\ref{subsec:differentiability} do indeed allow for efficient learning}
    \begin{tabularx}{\linewidth}{|C|C|}
        \hline
        Attention Method & $\ell_2$ loss \\
        \hline
        PSAL 1 & 0.00832 \\
        PSAL 3 & 0.00228 \\
        PSAL 7 & 0.00234 \\
        PSAL 15 & 0.00237 \\
        PSAL 31 & 0.00232 \\
        PSAL1 Aggreg & 0.001939 \\
        \hline
    \end{tabularx}
    \label{tab:metricsDifferentiability}
\end{table}

Figure~\ref{fig:reconstruction} shows the results of the image reconstruction task. We observe that not only does PSAL maintain good reconstruction, it in fact performs better than FA, which is not able to reconstruct fine details (see the zooms in the red squares on the bottom right of the images). This is a result of the stride induced by strong memory requirements. On the other hand, PSAL reconstruction has crisp details. This ability of PSAL to recover details is interesting when attention layers are used for tasks such as single image super-resolution~\citep{parmar_image_2018}.
More reconstruction results can be found in the appendix.

\begin{figure*}[!t]
\centering
\begin{subfigure}[t]{0.32\textwidth}
    \centering
    \includegraphics[width=\textwidth]{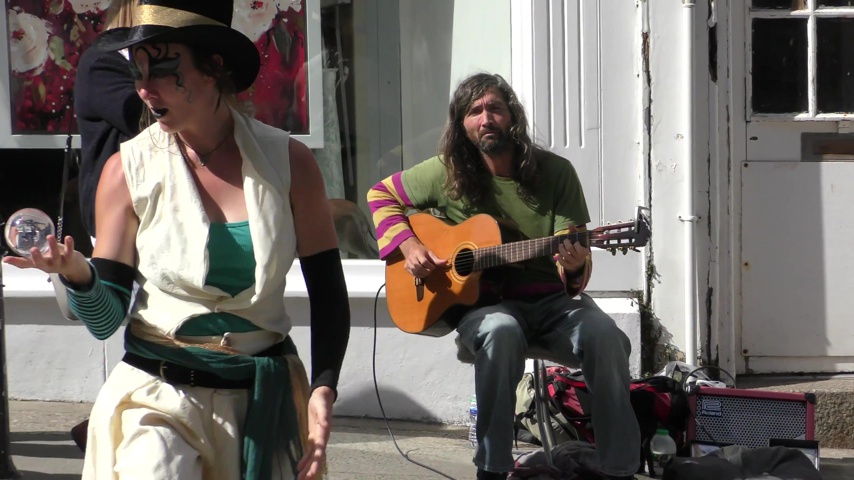}
    \caption{Ground truth}
\end{subfigure}
\hfill
\begin{subfigure}[t]{0.32\textwidth}
    \centering
    \includegraphics[width=\textwidth]{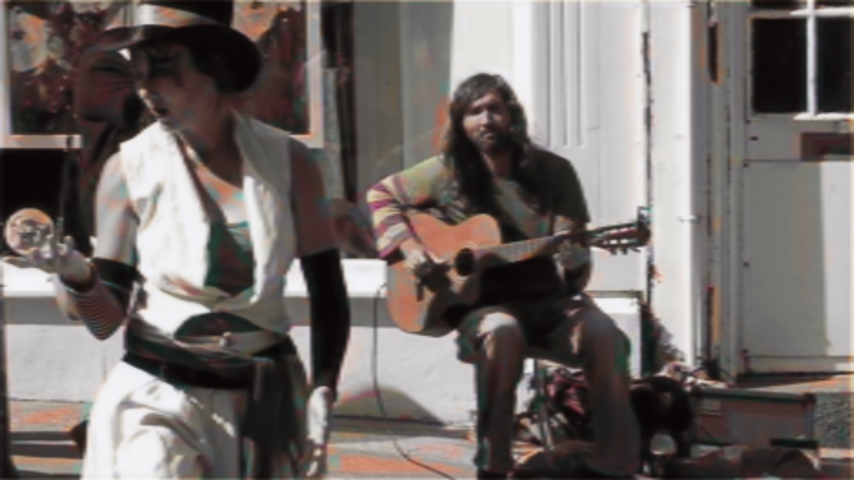}
    \caption{Full Attention}
\end{subfigure}
\hfill
\begin{subfigure}[t]{0.32\textwidth}
    \centering
    \includegraphics[width=\textwidth]{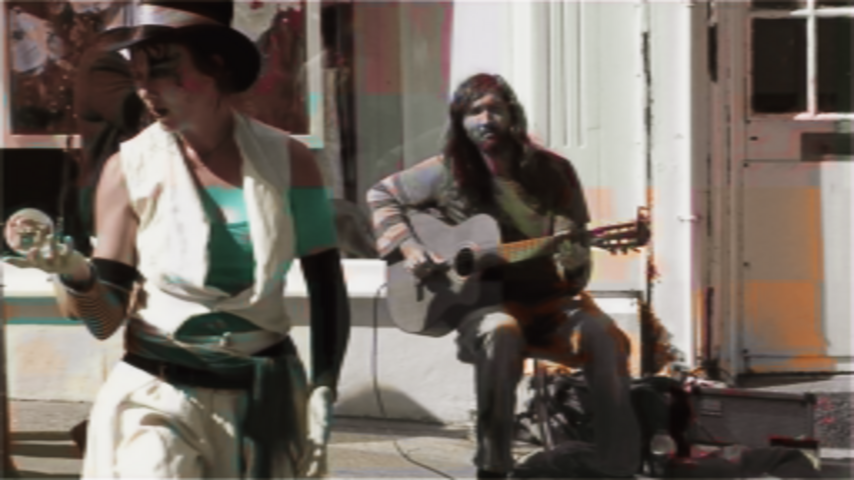}
    \caption{Local Attention}
\end{subfigure}

\begin{subfigure}[t]{0.32\textwidth}
    \centering
    \includegraphics[width=\textwidth]{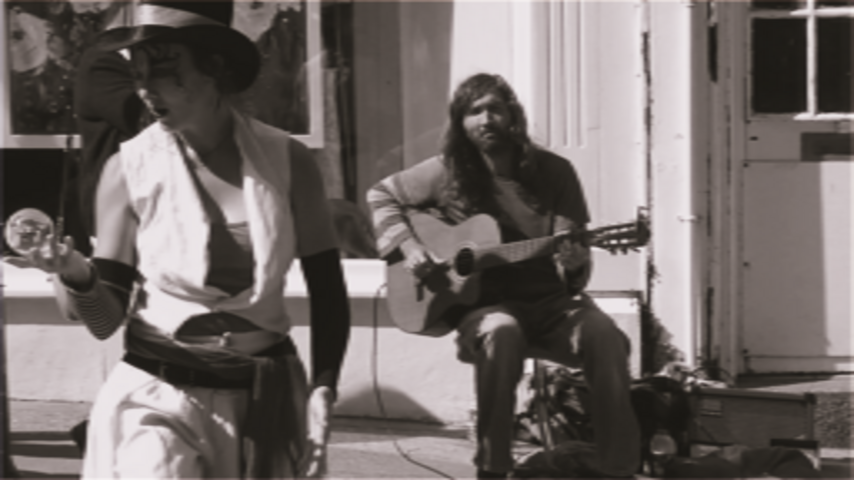}
    \caption{Performer}
\end{subfigure}
\hfill
\begin{subfigure}[t]{0.32\textwidth}
    \centering
    \includegraphics[width=\textwidth]{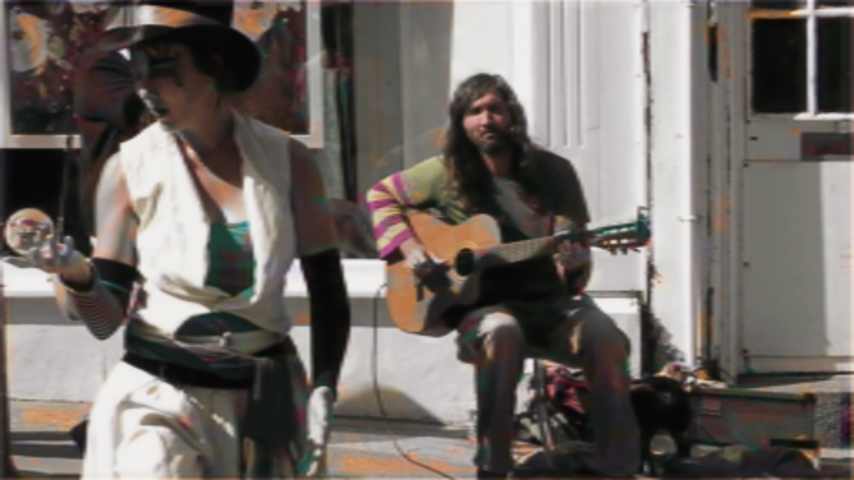}
    \caption{PSAL 3}
\end{subfigure}
\hfill
\begin{subfigure}[t]{0.32\textwidth}
    \centering
    \includegraphics[width=\textwidth]{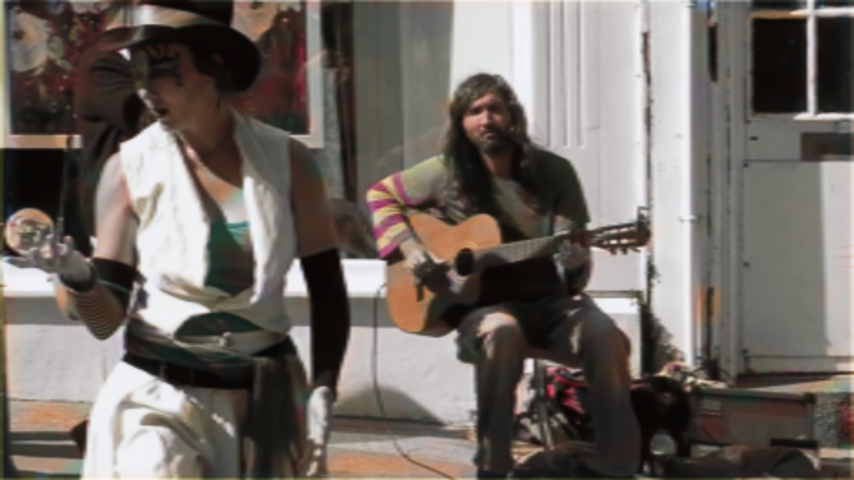}
    \caption{PSAL Aggreg.}
\end{subfigure}

\caption{Results on the colorization task. PSAL 3 and PSAL Aggreg. have good results despite some wrong matches (gray skin). }

\label{fig:colorizationImages}
\end{figure*}

\begin{figure*}[]\centering
\begin{tabular}{cc}
    \includegraphics[width=0.45\textwidth]{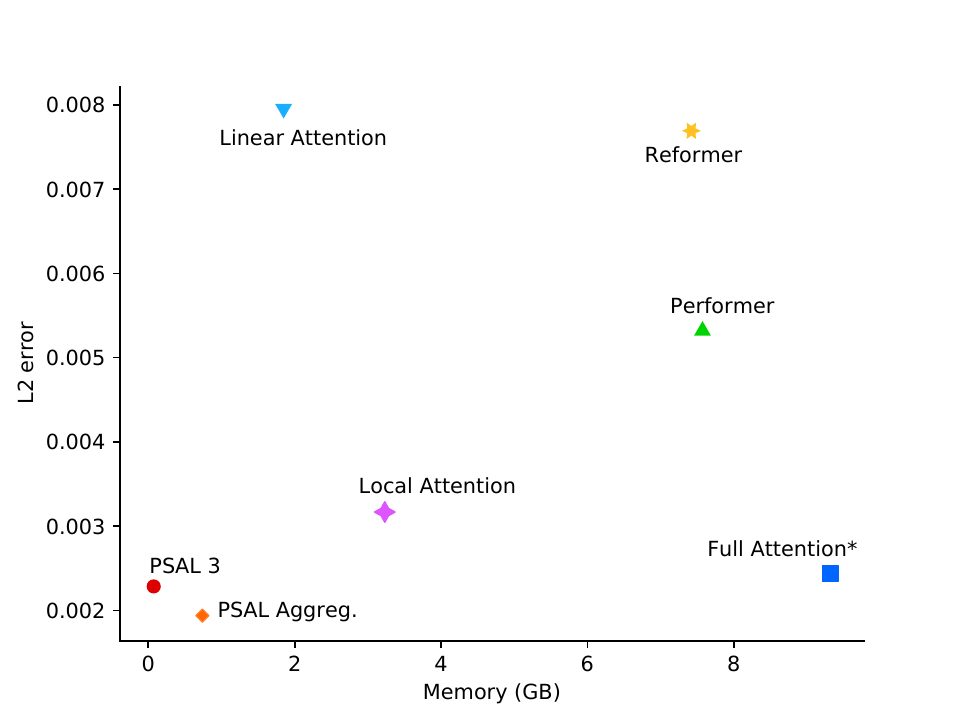} &
    \includegraphics[width=0.45\textwidth]{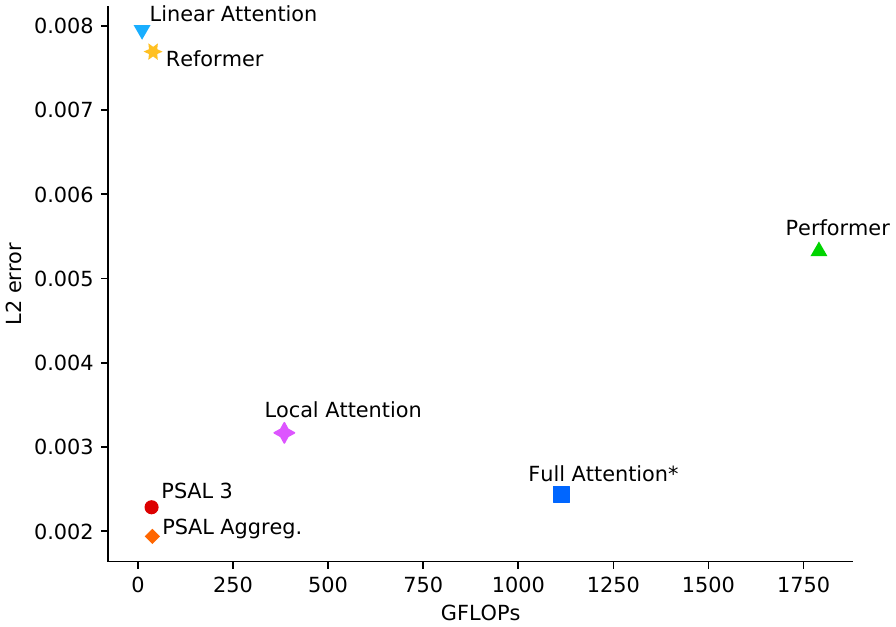}
\end{tabular}
\caption{Performance vs computation constraints (memory and GFLOPs) in the colorization task. Full Attention performs well at the cost of high memory and many GFLOPs. Local Attention is an efficient approximation of Full Attention with a limited drop in performance. Performer does not perform well. PSAL 3 and PSAL Aggreg. have better performance, largely reduced memory usage, and require less FLOPs than alternatives. $^*$Full Attention still requires a subsampling step to fit into GPU memory.}
\label{fig:memory_efficiency}
\end{figure*}

\subsection{Colorization}
\label{sec:colorization}

We evaluate PSAL on the task of guided image colorization. Given a grayscale image and a colorful reference image, we train a network to recover the color information. We use as simple a network as possible, to isolate the contribution of the attention layer. This architecture can be seen in Figure~\ref{fig:colorization}. Because we are using similar images, but not identical, attention is a crucial component to identify the best regions from which to copy the colors.

We compare our results with the other attention models by keeping the same architecture, but changing only the attention layer. We use images of size 256x256 which are large by feature map standards but low resolution in the modern context.
If we plot the $\ell_2$ error as a function of the memory usage or number of Floating Point OPerations (FLOPs), (Figure~\ref{fig:memory_efficiency}), it is clear that PSAL does not trade performance for memory or computations, and performs favorably against all other considered methods.
Full Attention has close results but is limited by memory constraints: to fit into memory a subsampling step (of stride $3$) is necessary which limits the set of keys considered.
For Local Attention~\citep{parmar_image_2018}, the model performs well for frames with little or no displacement but degrades abruptly when distant neighbors are required.
PSAL 3 reaches good results but may sometimes match patches incorrectly.
PSAL Aggreg. helps with this aspect, the aggregation step helps smoothing out irregularities.
As in the case of image reconstruction (Section~\ref{subsec:image_reconstruction}), the choice of either $\ell_2$ distance or dot product when comparing patches has a significant impact on performance. Our method is compatible with any metric.
For FA and Local Attention~\citep{parmar_image_2018}, we replaced the dot product with a $\ell_2$ operation.
For the Performer, which is based on softmax approximations of the dot product, we left them as described by their authors, since there is no obvious way to adapt their algorithm to the $\ell_2$ case. They indeed produce poorer results generally.
Reformer is very similar to our solution by using LSH to efficiently identify the nearest neighbors. For cross-attention, the proposed implementation of \citet{kitaev_reformer_2020} is far from perfect as it may create buckets containing only keys or queries, hence it has difficulty matching neighbors together.
We also compare our method to a linear approximation of the attention : Linear Attention~\citep{katharopoulos_et_al_2020}. In this case, we observe that non linearity is needed and the approximation is thus severely hurting the performance.

Figure~\ref{fig:colorizationImages} shows the visual results. We observe that FA and Performer produce faded results, while Local Attention has visible square artifacts because of local windows. PSAL Aggreg. produces the best results in this case. Further results are available in the supplementary material \ref{fig:colorizationImages2}.

\subsubsection{Training details}
The colorization network is minimal, only containing: a single layer of 3x3 convolutions (16 output channels), the attention layer, and another 3x3 convolution (3 output channels).
The goal is to put the emphasis on the attention layer and not on the problem of colorization, which is  complex with an extensive literature.
We add a residual connection so that the attention layer only has to provide color information.

Training is done on DAVIS dataset~\citep{perazzi_benchmark_2016}: train + test-dev. Testing is done on test-challenge.
Images are resized to 256x256. The patch size is set to 7.
We use the Adam optimizer with a learning rate of 0.001 for 200k iterations and a batch size of 1.

\subsubsection{$\ell_2$ vs dot product}
\label{subsec:l2}

For image colorization, we observed a large discrepancy between methods employing an $\ell_2$ distance and a dot product  similarity measure in the attention \eg $s(q,k) = \langle q, k \rangle$ vs. $s(q,k) = -\lVert q - k \rVert_2^2$.
Indeed, all methods using an $\ell_2$ distance (Full Attention, Local Attention, PSAL) are performing better than the methods based on dot product (Performer, Reformer, Linear Attention) in Figure~\ref{fig:memory_efficiency}.
Table~\ref{tab:l2} shows this gap for the Full Attention layer for the same hyper parameters.

We hypothesize that in this bare-bones experiment without additional layers and especially normalization layers, the dot product is not well adapted.

\begin{table}[]
    \centering
    \caption{$\ell_2$ vs dot product colorization results when using Full Attention. The choice of similarity metric is very important for the task. We see that $\ell_2$ performs much better.}
    \begin{tabular}{|c|c|}
        \hline
        Similarity function & $\ell_2$ loss \\
        \hline
        Dot product $s(q,k) = \langle q, k \rangle$ & 0.0064\\
        $\ell_2$ distance  $s(q,k) = -\lVert q - k \rVert_2^2$ &  0.0024\\
        \hline
    \end{tabular}
    \label{tab:l2}
\end{table}

\subsubsection{Differentiability and neighbors}
\label{sec:colorizationDifferentiability}

We now show experimental evidence that our proposed strategies for PSAL's differentiability are effective and are crucial for end-to-end training. Looking at Table~\ref{tab:metricsDifferentiability}, we see a large performance gap between PSAL-1 and PSAL-3/PSAL Aggreg. PSAL-1 is indeed not fully differentiable with respect to all its parameters and cannot learn the projection matrices $Q$ and $K$. The attention does not help in this case and performances are poor.
The performance does not improve beyond $k \geq 3$.
PSAL with aggregration uses more neighbors and aggregates values differently than PSAL-$k$.

\begin{table*}[]
    \centering
    \caption{Average inpainting metrics on Places2 validation set. Quantitative measures show similar performance using our layer with reduced memory requirements. $^*$ retrained.}
    \begin{tabular}{|c|c|c|c|c|c|}
        \hline
        Attention Method & $\ell_1$ loss $\downarrow$ & $\ell_2$ loss $\downarrow$ & PSNR (dB) $\uparrow$ & TV loss $\downarrow$ & SSIM $\uparrow$\\
        \hline
        ContextualAttention$^*$~\citep{yu_generative_2018} & 11.8\% & 3.6\% & 16.4 & \textbf{6.6\%} & 53.7 \\
        PSAL (ours) & \textbf{11.6\%} & \textbf{3.6\%} & \textbf{16.6} & 6.9\% & \textbf{54.1} \\
        \hline
    \end{tabular}
    \label{tab:metricsInpainting}
\end{table*}

PSAL-3, on the other hand, employs a 3-NN PSAL layer which makes optimization possible. Similarly, PSAL Aggreg. performs extremely well. This experiment confirms that our two approaches for making PSAL differentiable are effective.
This also shows that the adaptation of PatchMatch to feature maps is not as straightforward as it looks.

\subsection{Image inpainting}
\label{sec:imageInpainting}

A prominent field of application of attention models is image inpainting. This is the process of automatically filling in unknown or damaged regions in an image.
Deep learning inpainting algorithms are able to inpaint semantic objects, such as positioning an eye in a face. However, one of the blindspots of such approaches to image inpainting is the correct reconstruction of textures and fine details.
It turns out that this problem is, in turn, well addressed by using patch-based approaches~\citep{criminisi_region_2004,wexler_space-time_2007}. Thus, the ideal inpainting method would be able to unite the strengths of both deep learning and patch-based approaches in a single algorithm. This has motivated the introduction of attention layers in deep inpainting networks.

Recently, \citet{yu_generative_2018} have introduced an attention layer with great success in their inpainting network, which we refer to as ContextualAttention (CA).
After a first coarse inpainting, the image is refined flowing into 2 different branches: a fully convolutional network, and an attention-based network.
The outputs are then merged.
At its core, CA is a Full Attention layer based on 3x3 patches in feature maps.
Unfortunately, even with mid-level features, the remaining spatial size is too large to avoid a memory overflow especially during training.
Yu \etal limit the number of patches to be computed using a downsampling scheme. Once again, this illustrates the practical need for an attention layer which scales to large images.
Given the setting, the proposed PSAL is a good fit in the network.

We directly replace the FA layer with our PSAL 3. 
We use a patch size of 7 for PSAL, which is equivalent to a patch size of 3 plus a downsampling with a factor of 2 as used by CA.
The quantitative results shows no significant differences between PSAL and ContextualAttention for  different metrics (Table~\ref{tab:metricsInpainting}).
We also include the Total Variation (TV) loss which measures the smoothness of the inpainted image and does not depend on the ground truth: 

$$
\sum_{i,j} \lvert x_{i+1} - x_{i} \rvert + \lvert x_{j+1} - x_{j} \rvert
$$

This confirms that the PSAL can indeed replace the CA layer, with no quality loss, but with a great reduction in memory requirements.

Note that we do not need to compare with other attention approaches, since our goal here is not to improve the quality of inpainting, but rather to show that our PSAL can be easily inserted into any existing architecture, without any loss in quality, while greatly reducing memory requirements (which we showed in Table~\ref{tab:benchmark}).

\begin{figure}[t]
    \centering
    \includegraphics[width=0.5\textwidth]{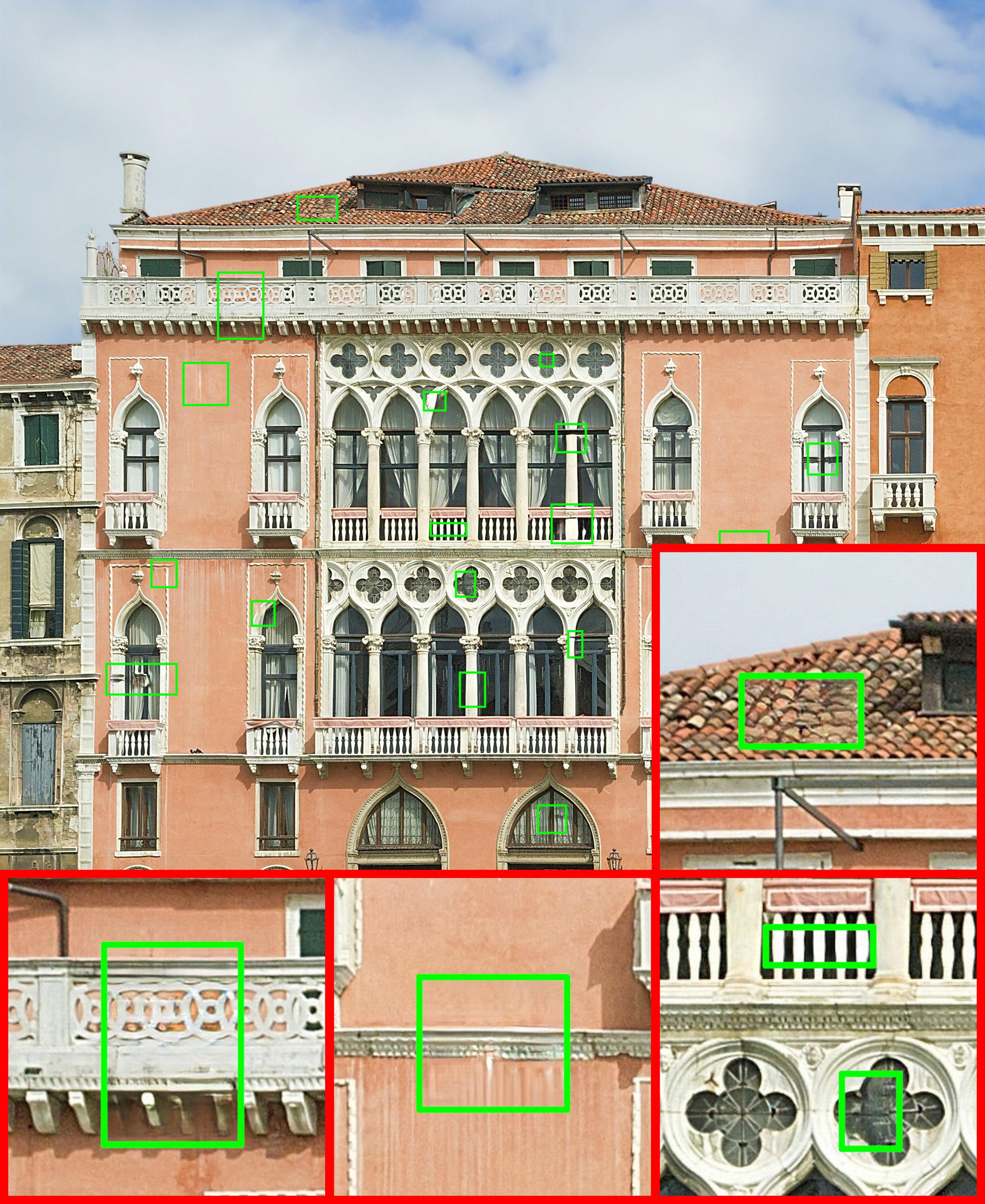}
    \vskip -0.1in
    \caption{A 2700x3300 image inpainted using PSAL. The initial occlusions are outlined in green. We show the full page version in the supplementary material (\ref{fig:hires1}). Original picture by Didier Descouens - Licensed under CC BY-SA 4.0}
    \label{fig:high_res_inpainting_paper}
\end{figure}

\subsubsection{Inpainting results}
\label{sec:inpainting_appendix}

\begin{figure*}
\centering
\begin{subfigure}[t]{0.2\textwidth}
    \centering
    \includegraphics[width=\textwidth]{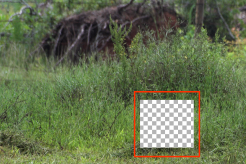}
    \caption{Ground truth}
\end{subfigure}
\begin{subfigure}[t]{0.13\textwidth}
    \centering
    \includegraphics[width=\textwidth]{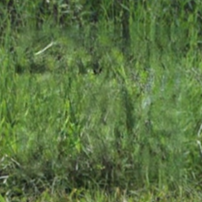}
    \caption{ContextualAttention}
\end{subfigure}
\begin{subfigure}[t]{0.13\textwidth}
    \centering
    \includegraphics[width=\textwidth]{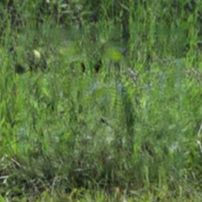}
    \caption{PSAL}
\end{subfigure}
\hfill
\begin{subfigure}[t]{0.2\textwidth}
    \centering
    \includegraphics[width=\textwidth]{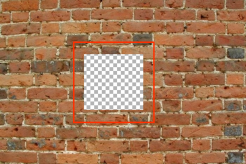}
    \caption{Ground truth}
\end{subfigure}
\begin{subfigure}[t]{0.13\textwidth}
    \centering
    \includegraphics[width=\textwidth]{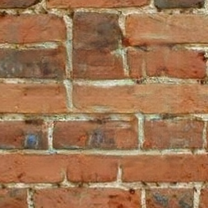}
    \caption{ContextualAttention}
\end{subfigure}
\begin{subfigure}[t]{0.13\textwidth}
    \centering
    \includegraphics[width=\textwidth]{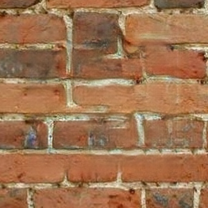}
    \caption{PSAL}
\end{subfigure}

\begin{subfigure}[t]{0.2\textwidth}
    \centering
    \includegraphics[width=\textwidth]{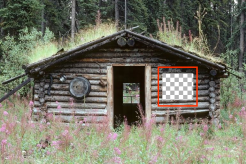}
    \caption{Ground truth}
\end{subfigure}
\begin{subfigure}[t]{0.13\textwidth}
    \centering
    \includegraphics[width=\textwidth]{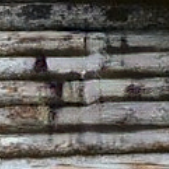}
    \caption{ContextualAttention}
\end{subfigure}
\begin{subfigure}[t]{0.13\textwidth}
    \centering
    \includegraphics[width=\textwidth]{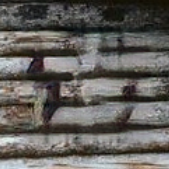}
    \caption{PSAL}
\end{subfigure}
\hfill
\begin{subfigure}[t]{0.2\textwidth}
    \centering
    \includegraphics[width=\textwidth]{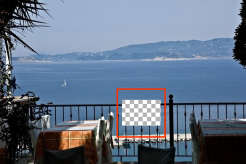}
    \caption{Ground truth}
\end{subfigure}
\begin{subfigure}[t]{0.13\textwidth}
    \centering
    \includegraphics[width=\textwidth]{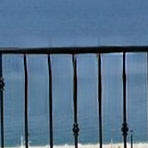}
    \caption{ContextualAttention}
\end{subfigure}
\begin{subfigure}[t]{0.13\textwidth}
    \centering
    \includegraphics[width=\textwidth]{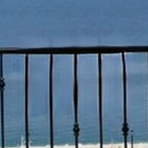}
    \caption{PSAL}
\end{subfigure}

\caption{Here, we show the results of the ContextualAttention~\citep{yu_generative_2018} and our version of the algorithm where the FA layer is replaced with a PSAL. We observe that the results are of very similar quality, validating the direct and straightforward replacement of FA with PSAL.}
\label{fig:context}
\end{figure*}

For completeness, we produce visual comparisons of inpainting results.
Figure~\ref{fig:context} shows very similar results between our method and the one from \cite{yu_generative_2018}. This is exactly what we aimed for, since we want to achieve the same inpainting results with our more memory efficient PSAL. This is reflected by close quantitative inpainting metrics.

\subsubsection{Training}

We retrain Contextual Attention~\citep{yu_generative_2018} using the same hyper-parameters than the authors and an implementation by Du Ang {\small\url{https://github.com/daa233/generative-inpainting-pytorch}}.
Specifically, we train our networks for 800k iterations with a batch size of 16 on Places2.

For PSAL, we train for the same number of iterations. We use a patch size of 7, remove the downsampling step in the attention, reconstruct only from the central pixel, and use 5 iterations of PatchMatch.

\subsection{High resolution inpainting}

Finally, in Figure~\ref{fig:high_res_inpainting_paper} we show that with PSAL we can inpaint high resolution images, which is the initial goal that motivates this work (extending attention layers to larger resolution images).
The low memory requirements make it possible to handle images of resolution up to 3300x3300 on a NVIDIA GTX 1080 Ti with 11 GB of memory.
Processing such a large image using the FA layer without subsampling would require more than 1000 GB of memory.

The images are processed at their native resolutions without memory-saving tricks.
The network is still limited by its receptive field which makes it only possible to fill small holes (128x128).
Note that the maximum size of the occlusions/holes (128$\times$128) is imposed by the network architecture of ContextualAttention. We could certainly create another architecture using PSAL for larger occlusion sizes, but to be fair to the original work we kept the same sizes.

\subsection{Super-Resolution}
\label{sec:super-resolution}

Attention can also be very useful for the super-resolution task~\citep{liu_non-local_2018}. Patch recurrence in natural images at different scales have been identified as a strong prior for super resolution, providing rich information~\citep{glasnerSuperresolutionSingleImage2009,michaeliBlindDeblurringUsing2014}. Recent adaptations to attention with In-Scale attention and Cross-Scale Attention by \citet{meiImageSuperResolutionCrossScale2020} confirm these findings for deep learning.

Recurring patches naturally occur at long range due to perspective, thus limiting the usefulness of local or restrained attention. Looking at the work of Mei~\etal, we propose to replace the Full Attention layer which they use in the Cross-Scale branch by PSAL.

Our results (Table~\ref{tab:super-resolution}) on the Urban 100 dataset for different zoom factors indicate that we can use PSAL to efficiently approximate the Cross-Scale attention which benefits from long range dependencies. The In-Scale attention branch is more efficiently approximated with a local attention. We test our hypothesis by keeping the same weights and changing only the Cross-Scale attention layer.
Similar performance is reached using only a fraction of the memory (Table~\ref{tab:super-resolution}), this is of high interest for handling images with repetitive structures at a distance \ie at high resolution.

\begin{table}[]
    \centering
    \caption{For single-image super-resolution, Cross-Scale attention~\citep{meiImageSuperResolutionCrossScale2020} can be efficiently approximated with PSAL as indicated by similar PSNR scores on the Urban 100 dataset.}
    \begin{tabular}{|c|c|c|c|}
        \hline
        Attention Method & Zoom x2 & Zoom x3 & Zoom x4 \\
        \hline
        Cross-Scale Attention & 33.383 & 29.123 & 27.288\\
        PSAL & 33.375 & 29.112 & 27.184\\
        \hline
    \end{tabular}
    \label{tab:super-resolution}
\end{table}

Early works on patch recurrence~\cite{glasnerSuperresolutionSingleImage2009,michaeliBlindDeblurringUsing2014} showed that the benefits of Cross-Scale are most spectacular at very large resolutions and zoom levels, which cannot be achieved by full attention implementations such as the one in~\cite{meiImageSuperResolutionCrossScale2020} due to GPU memory limitations. By replacing this implementation with PSAL in the Cross-Scale branch we expect to be able to retrain the architecture for much larger zoom levels and image sizes, thus showing the full potential of patch recurrence. This will be the subject of future research.

\subsection{Pretrained networks}
\label{sec:pretrained}

One advantage of our attention layer is that it can replace the attention layers in already trained models.
More generally, because PSAL-$k$ is an approximation of Full Attention, we can replace attention layers, keeping similar results at a fraction of the original computational cost.
This is particularly useful when performing inference on large inputs.

We show this effect on the colorization task using PSAL-1 as a replacement for other attention layers at inference time.
While the parameters of PSAL-1 are not all trainable, we can still use it if we do not need to train the network. PSAL-1 is still a fast, light, and good approximation of Full Attention or other PSAL-$k$ layers.
In the colorization task, we trained and froze several different networks, and then we switched attention layers to PSAL-1. We observed very good results with this approach.
Table~\ref{tab:pretrained} shows for instance that Full Attention can be approximated with no drop in performance but with 40x less computations and 225x less memory.

\begin{table*}[]
    \centering
    \caption{Colorization performance when using weights from pretrained models but replacing the attention layer with PSAL-1 at test time. PSAL-1 can replace the attention layers in trained networks with similar performance but a significant reduction in FLOPS and memory usage.}
    \vskip 0.15in
    \begin{tabular}{|c|c|c|c|c|}
        \hline
        Attention during training & Attention during test & $\ell_2$ loss & Test GFLOPs & Test Memory (GB) \\
        \hline
        PSAL 1 & PSAL 1 & 0.0083 & 30 & 0.05\\
        \hline
        PSAL Aggreg & PSAL Aggreg. & 0.0019 & 37 & 0.74\\
        PSAL Aggreg. & PSAL 1 & 0.0023 & 30 & 0.05\\
        \hline
        PSAL 3 & PSAL 3 & 0.0023 & 36 & 0.08\\
        PSAL 3 & PSAL 1 & 0.0023 & 30 & 0.05\\
        \hline
        Full Attention & Full Attention & 0.0024 & 1173 & 9.32\\
        Full Attention & PSAL 1 & 0.0024 & 30 & 0.045\\
        \hline
        Local Attention & Local Attention & 0.0032 & 385 & 3.23\\
        Local Attention & PSAL 1 & 0.0035 & 30 & 0.045\\
        \hline
    \end{tabular}
    \label{tab:pretrained}
\end{table*}

\section{Conclusion}

In this work, we have presented PSAL, an efficient patch-based stochastic attention layer that is not limited by GPU memory.
We have showed that our layer gives a much lighter memory load, scaling to very high resolution images.
This makes the processing of high resolution images possible with deep networks using attention, without any of the customary tricks currently used (subsampling, etc).
Furthermore, new network architectures using attention mechanisms on low level features are now conceivable. We have demonstrated the use of PSAL in several tasks, showing in particular that high resolution image inpainting is achievable with PSAL.

We plan to continue this work by applying the proposed attention layer to other image editing tasks, and in particular to be able to process videos, which is not achievable at this point in time by classical attention architectures, due to the memory constraints addressed in the present work. PSAL could also possibly be employed for the most recent, well-performing, neural networks involved in restoration tasks, such as diffusion models.

\section*{Acknowledgements}
The authors acknowledge support from the French Research Agency through the PostProdLEAP project (ANR-19-
CE23-0027-01). Nicolas Cherel is supported by the Institut Mines-Télécom, Fondation Mines-Télécom, and l’Institut Carnot TSN through the "Futur et Ruptures" grant.

\bibliography{references}
\bibliographystyle{model2-names}

\newpage
\appendix
\onecolumn

\section{Reconstruction}
\begin{figure*}
    \centering
\begin{subfigure}[t]{0.45\textwidth}
    \includegraphics[width=\textwidth]{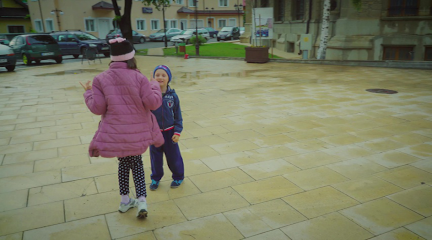}
    \caption{Ground Truth}
\end{subfigure}
    \begin{subfigure}[t]{0.45\textwidth}
    \includegraphics[width=\textwidth,height=0.56\textwidth]{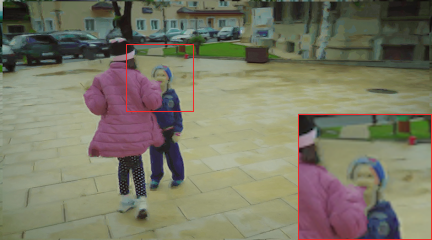}
    \caption{PSAL-3 result}
\end{subfigure}
    \caption{Reconstruction failure case with PSAL}
    \label{fig:failure_rec}
\end{figure*}

To complete Section~\ref{subsec:image_reconstruction}, we present additional reconstruction results comparing PSAL 3, PSAL Aggreg. and Full Attention (stride 10), see Figure~\ref{fig:colorizationImages2}. A failure case is shown in Figure~\ref{fig:failure_rec}.

\begin{figure*}[]\centering
\begin{tabular}{ccc}
Full Attention & PSAL 3 & PSAL Aggreg. \\
\includegraphics[width=0.30\textwidth,height=0.168\textwidth]{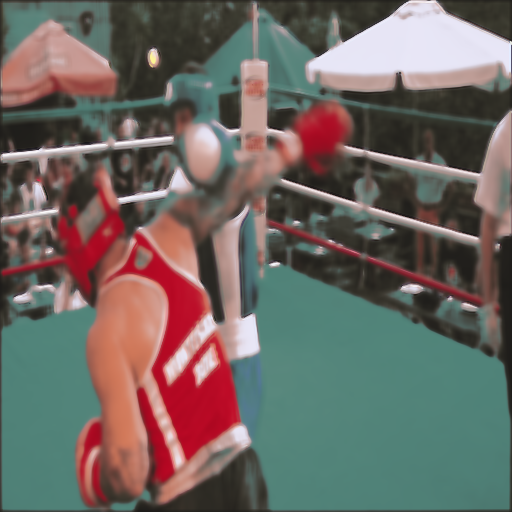} &
\includegraphics[width=0.30\textwidth,height=0.168\textwidth]{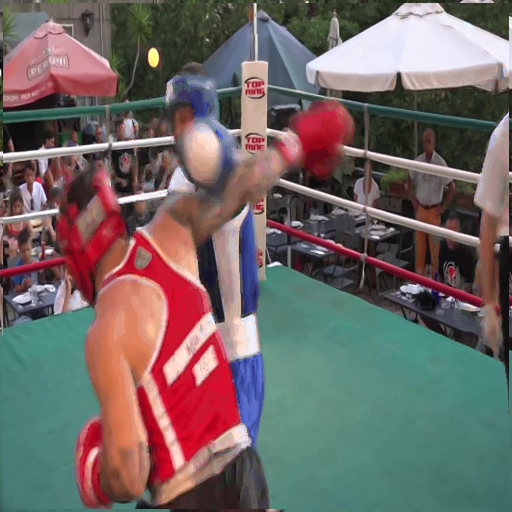} &
\includegraphics[width=0.30\textwidth,height=0.168\textwidth]{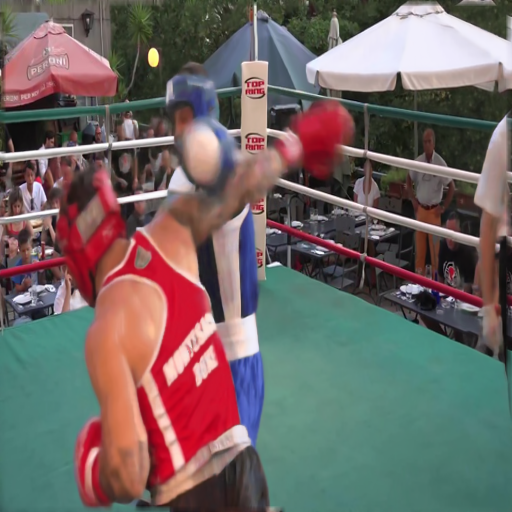} \\
\includegraphics[width=0.30\textwidth,height=0.168\textwidth]{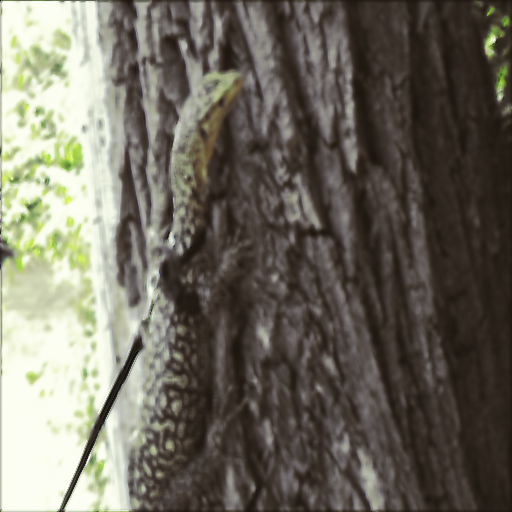} &
\includegraphics[width=0.30\textwidth,height=0.168\textwidth]{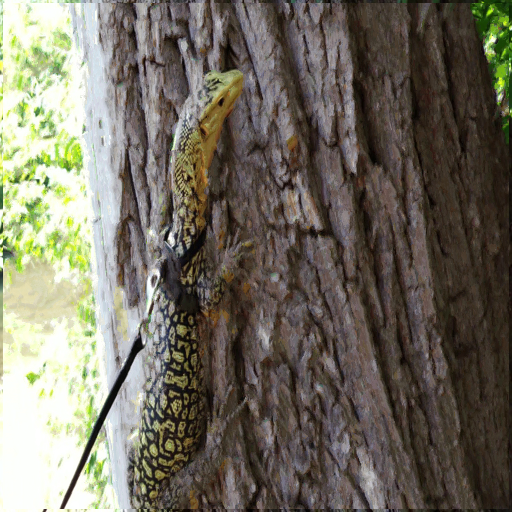} &
\includegraphics[width=0.30\textwidth,height=0.168\textwidth]{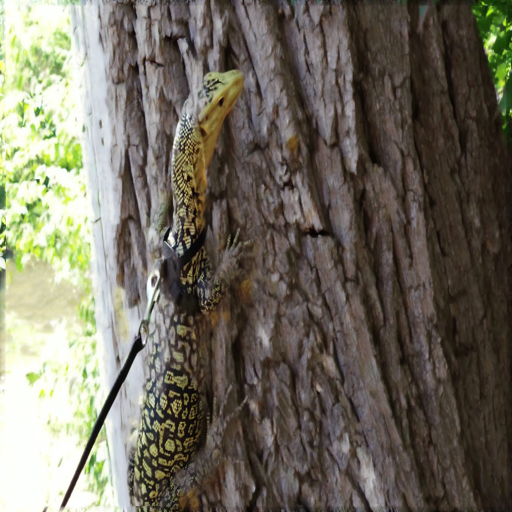} \\
\end{tabular}
\caption{More results on the reconstruction task. Best seen with zoom-in.}
\label{fig:colorizationImages2}
\end{figure*}

\section{Colorization}

\begin{figure*}[]\centering

\begin{subfigure}[t]{0.32\textwidth}
    \centering
    \includegraphics[width=\textwidth,height=0.56\textwidth]{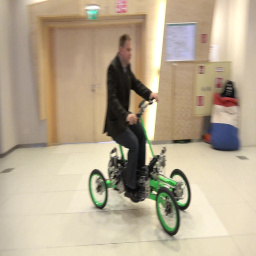}
    \caption{Ground truth}
\end{subfigure}
\hfill
\begin{subfigure}[t]{0.32\textwidth}
    \centering
    \includegraphics[width=\textwidth,height=0.56\textwidth]{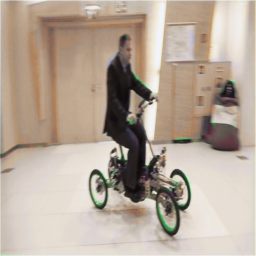}
    \caption{Full Attention}
\end{subfigure}
\hfill
\begin{subfigure}[t]{0.32\textwidth}
    \centering
    \includegraphics[width=\textwidth,height=0.56\textwidth]{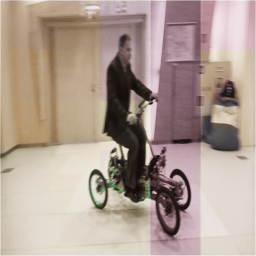}
    \caption{Local Attention}
\end{subfigure}

\begin{subfigure}[t]{0.32\textwidth}
    \centering
    \includegraphics[width=\textwidth,height=0.56\textwidth]{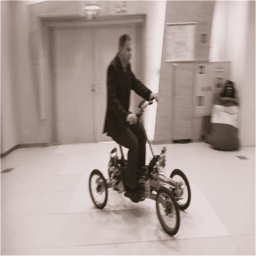}
    \caption{Performer}
\end{subfigure}
\hfill
\begin{subfigure}[t]{0.32\textwidth}
    \centering
    \includegraphics[width=\textwidth,height=0.56\textwidth]{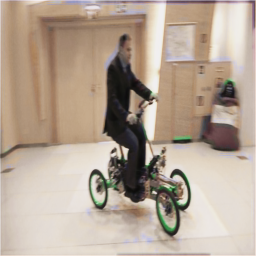}
    \caption{PSAL 3}
\end{subfigure}
\hfill
\begin{subfigure}[t]{0.32\textwidth}
    \centering
    \includegraphics[width=\textwidth,height=0.56\textwidth]{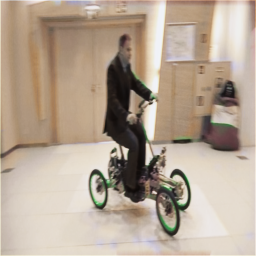}
    \caption{PSAL Aggreg.}
\end{subfigure}

\vspace{1cm}

\begin{subfigure}[t]{0.32\textwidth}
    \centering
    \includegraphics[width=\textwidth,height=0.56\textwidth]{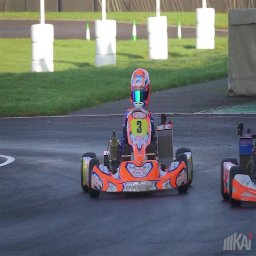}
    \caption{Ground truth}
\end{subfigure}
\hfill
\begin{subfigure}[t]{0.32\textwidth}
    \centering
    \includegraphics[width=\textwidth,height=0.56\textwidth]{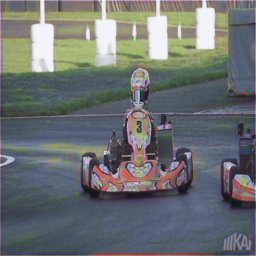}
    \caption{Full Attention}
\end{subfigure}
\hfill
\begin{subfigure}[t]{0.32\textwidth}
    \centering
    \includegraphics[width=\textwidth,height=0.56\textwidth]{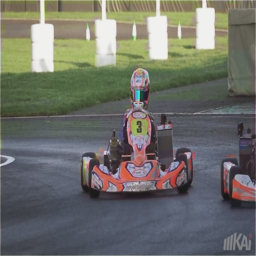}
    \caption{Local Attention}
\end{subfigure}

\begin{subfigure}[t]{0.32\textwidth}
    \centering
    \includegraphics[width=\textwidth,height=0.56\textwidth]{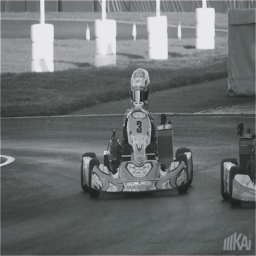}
    \caption{Performer}
\end{subfigure}
\hfill
\begin{subfigure}[t]{0.32\textwidth}
    \centering
    \includegraphics[width=\textwidth,height=0.56\textwidth]{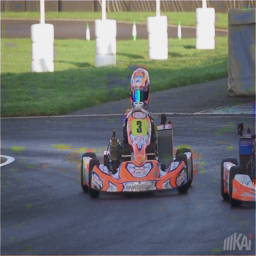}
    \caption{PSAL 3}
\end{subfigure}
\hfill
\begin{subfigure}[t]{0.32\textwidth}
    \centering
    \includegraphics[width=\textwidth,height=0.56\textwidth]{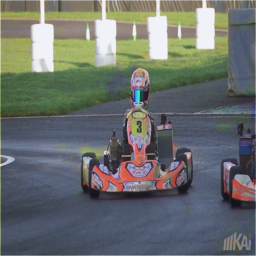}
    \caption{PSAL Aggreg.}
\end{subfigure}

\caption{More results on the colorization task. Performer and Full Attention produce bland results. Because of large displacements between frames local attention is not enough to recover the true colors resulting in artifacts. PSAL 3 and PSAL Aggreg. have good results despite some wrong matches.}
\label{fig:colorizationImages2}
\end{figure*}

\subsection{Why does Performer perform so badly on our colorization benchmark?}

We split this question into two subsequent questions: 1) Why Performer does not successfully complete the task? 2) Why are Performer's memory and FLOPS worse than Full Attention despite linear complexity?

For the first question, we gave a hint in Section~\ref{subsec:l2}. For a simple network without normalization layers, the $\ell_2$ patch distance is a better metric than the dot product.
Indeed, in the case of unnormalized patches, the dot product is maximized by patches of large norm rather than similar patches. This has been observed for Full Attention and we hypothesize that the same conclusion holds for Performer.
We have not found an obvious adaptation of Performer to the $\ell_2$ metric.

For the second question, we used the recommended parameter, setting the dimension of the projection to $M=d \log d$.
In our case, the patch dimension $d$ is $7\times 7 \times 16 = 784$ and therefore $M=784 \log (784) = 5224$.
This explains the large memory usage: contrary to the other methods, each patch must be gathered and projected to this large space.

\subsection{Failure cases}
\begin{figure*}
    \centering
\begin{subfigure}[t]{0.45\textwidth}
    \includegraphics[width=\textwidth]{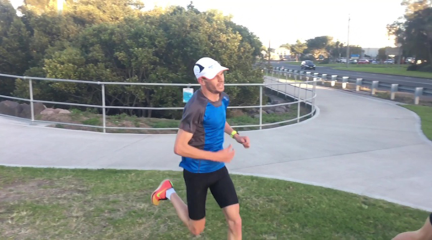}
    \caption{Ground Truth}
\end{subfigure}
    \begin{subfigure}[t]{0.45\textwidth}
    \includegraphics[width=\textwidth,height=0.56\textwidth]{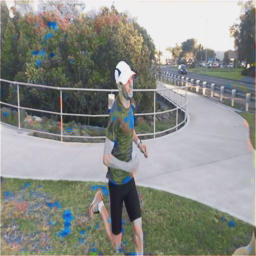}
    \caption{PSAL-3 result}
\end{subfigure}
    \caption{Colorization failure case with PSAL: the colors are wrongly matched}
    \label{fig:failure_col}
\end{figure*}

An example of a failure case for the colorization task is shown in Figure~\ref{fig:failure_col}.\\

\section{High resolution inpainting}

\begin{figure*}
    \centering
    \includegraphics[width=\textwidth]{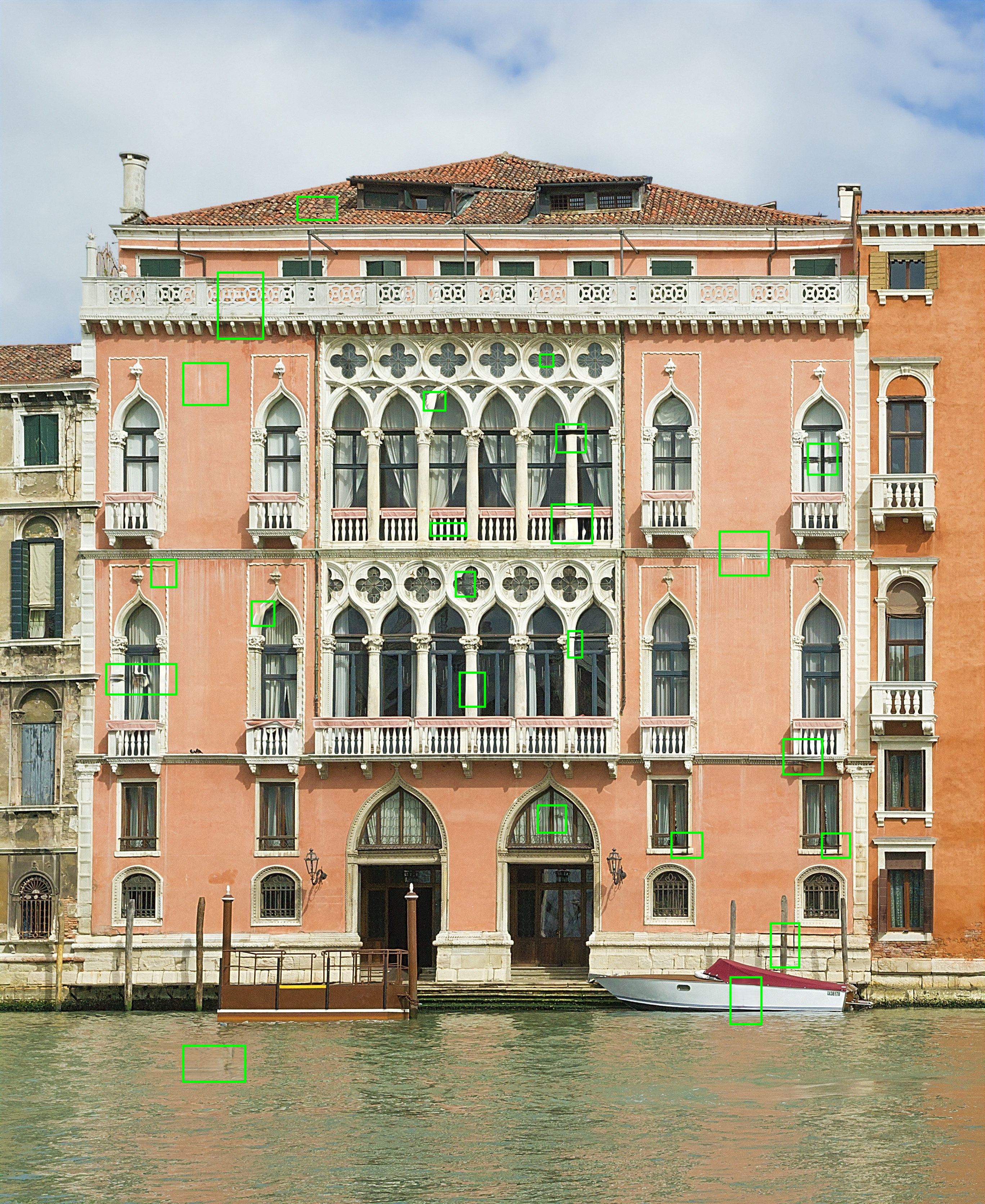}
    \caption{A 2700x3300 image inpainted using PSAL. The initial occlusions are outlined in green. Note that the maximum size of the occlusions/holes (128$\times$128) is imposed by the network architecture of ContextualAttention. We could certainly create another architecture using PSAL for larger occlusion sizes, but to be fair to the original work we kept the same sizes. Original picture by Didier Descouens - Licensed under CC BY-SA 4.0}
    \label{fig:hires1}
\end{figure*}

\begin{figure*}
    \centering
    \includegraphics[width=\textwidth]{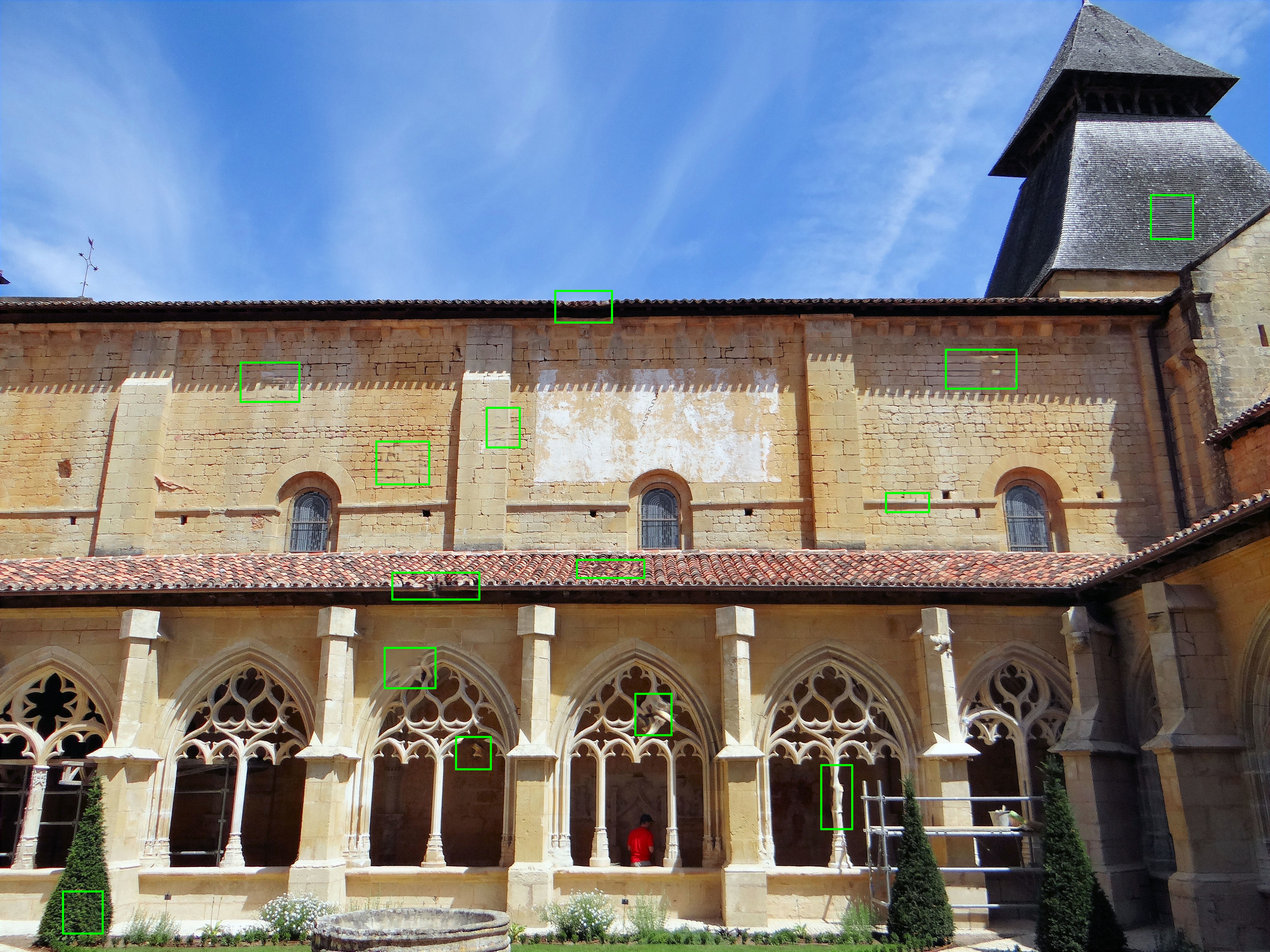}
    \caption{A 3600x2700 image inpainted using PSAL. The initial occlusions are outlined in green. Original picture by MOSSOT - Licensed under CC BY-SA 3.0}
    \label{fig:hires2}
\end{figure*}

We present further high resolution inpainting results.
Figure~\ref{fig:hires1} and Figure~\ref{fig:hires2} really highlight the completion of details and structures for which attention is important

\end{document}